\documentclass{article}

\usepackage{microtype}
\usepackage{graphicx}
\usepackage{booktabs} 

\usepackage{hyperref}

\usepackage[accepted]{icml2025}

\usepackage{amsmath}
\usepackage{amssymb}
\usepackage{mathtools}
\usepackage{amsthm}

\usepackage{multirow}
\usepackage{subcaption}

\usepackage[capitalize,noabbrev]{cleveref}

\theoremstyle{plain}

\theoremstyle{definition}

\theoremstyle{remark}

\usepackage[textsize=tiny]{todonotes}

\icmltitlerunning{Multi-Task Reward Learning from Human Ratings}

\begin{document}

\twocolumn[
\icmltitle{Multi-Task Reward Learning from Human Ratings}




\begin{icmlauthorlist}
\icmlauthor{Mingkang Wu}{yyy}
\icmlauthor{Devin White}{comp}
\icmlauthor{Evelyn Rose}{yyy}
\icmlauthor{Vernon Lawhern}{sch}
\icmlauthor{Nicholas R Waytowich}{sch}
\icmlauthor{Yongcan Cao}{yyy}
\end{icmlauthorlist}

\icmlaffiliation{yyy}{Department of Electrical and Computer Engineering, University of Texas at San Antonio, Texas, USA}
\icmlaffiliation{comp}{Army Educational Outreach Program, USA}
\icmlaffiliation{sch}{DEVCOM Army Research Lab, USA}

\icmlcorrespondingauthor{Mingkang Wu}{mingkang.wu@utsa.edu}

\icmlkeywords{Reinforcement Learning from Human Feedback, RLHF, RbRL, Rating-based Reinforcement Learning, AI Alignment, Reinforcement Learning, Reward Learning, Human Feedback, Multi-Task learning}

\vskip 0.3in
]



\printAffiliationsAndNotice{} 

\begin{abstract}
Reinforcement learning from human feedback (RLHF) has become a key factor in aligning model behavior with users' goals. However, while humans integrate multiple strategies when making decisions, current RLHF approaches often simplify this process by modeling human reasoning through isolated tasks such as classification or regression. In this paper, we propose a novel reinforcement learning (RL) method that mimics human decision-making by jointly considering multiple tasks. Specifically, we leverage human ratings in reward-free environments to infer a reward function, introducing learnable weights that balance the contributions of both classification and regression models. This design captures the inherent uncertainty in human decision-making and allows the model to adaptively emphasize different strategies. We conduct several experiments using synthetic human ratings to validate the effectiveness of the proposed approach. Results show that our method consistently outperforms existing rating-based RL methods, and in some cases, even surpasses traditional RL approaches. 
\end{abstract}

\section{Introduction}
\label{introduction}

Reinforcement learning (RL) has been developed and improved to solve complex tasks \textcolor{black}{via trial and error} \citep{sutton1998introduction}. 
\textcolor{black}{In standard RL, an agent, such as a robot or an autonomous vehicle, traverses through an environment gathering a reward at each step. The objective is to learn an optimal policy that maximizes the cumulative reward.}


RL differs from traditional deep learning methods as it relies heavily on the reward and the reward functions throughout training \citep{sutton1999reinforcement}. Typically, the reward reflects the explicit value of taking a certain action given a specific state. However, designing a proper reward function requires comprehensive understanding of the environment and can be difficult to cover all behaviors which lead to an optimal policy \citep{wu2024offline}. Existing environments which are used to benchmark RL algorithms, such as Atari, MuJoCo and DeepMind Control Suite \citep{mnih2013playing, todorov2012mujoco, tassa2018deepmindcontrolsuite}, have custom tailored reward functions designed to optimize a policy for a given objective. However, in real-world scenarios, such as transportation systems or drone navigation \citep{wu2023value}, it is challenging to design a reward function that accounts for all possible situations. For instance, a well-designed reward function should penalize collisions during obstacle avoidance while also encouraging path following. However, when an obstacle lies directly on the pre-planned path, the reward signals can conflict, making it difficult for the agent to determine the correct behavior. This conflict often results in suboptimal or unsafe decisions, underscoring the limitations of handcrafted reward functions and the need for learning-based approaches that can adapt to complex and dynamic environments.

A promising approach is to learn a reward function directly from human feedback. In this paradigm, users are typically shown short video segments or pairs of segments and asked to provide ratings or preferences \citep{christiano2017deep, white2024rating, rose2025performance, wu2025rbrl2}. These ratings or preferences would then be used to train a reward model that will be used to train the policy. Existing preference-based and rating-based reinforcement learning methods \citep{christiano2017deep, white2024rating} typically rely on customized classification models to predict human preferences or ratings for unseen agent behaviors. While such approaches have shown effectiveness when combined with human feedback, we argue that relying solely on classification models may not fully capture the nuanced human intuitions behind these preferences and ratings. To address this, we propose a method that frames reward learning as a multi-task prediction problem incorporating human input. Specifically, our approach leverages not only a classification model but also a regression model, allowing us to directly extract insights from human criteria, particularly through the use of ratings, when evaluating agent behaviors.

Specifically, in addition to incorporating a classification model to learn the reward function, we first map discrete human ratings to smoothed continuous values, which are treated as reward labels corresponding to the observed behaviors. These values serve as ground truth targets in a regression objective that minimizes the discrepancy between the predicted reward and the reward derived from human ratings. By considering this regression term, we propose a new multi-task reward learning framework, beyond the classification-based reward learning, that can better captures human intuition. However, it is challenging to assess the relative contribution of each objective during training, since human-provided ratings are inherently imprecise. To address this, we introduce learnable weights for each objective \citep{kendall2018multi}, allowing the model to dynamically adjust the confidence of the classification and regression components. When the confidence of one objective is higher, the model relies more heavily on that component at the current training step. Finally, we conduct several experiments to evaluate the effectiveness of the proposed method. The results demonstrate that the proposed new approach can yield significant improvements over the standard classification-based approach, and can match or even outperform the state-of-the-art RL method.

Here are a list of the main contribution of this paper. First, we propose a novel mapping strategy that transforms discrete human ratings into smoothed continuous values, which can be directly used as reward signals for RL agents. This allows the agent to update its policy based on rewards that more accurately reflect the underlying human intuition behind the ratings. 
Second, we introduce a learnable weighting mechanism that dynamically balances the contributions of classification and regression objectives during training. This adaptive approach enables the reward model to rely more heavily on the objective that better captures human feedback at a given stage of learning.
Third, we conduct extensive experiments across multiple environments to demonstrate that our proposed multi-task reward learning framework significantly outperforms prior classification-based methods.

\section{Related Work}
\label{related work}

RLHF has recently become a thriving area, widely used to guide both pre-trained and training-in-progress models towards behaviors that humans expect \citep{dai2023safe}. The problem we address focuses on learning entirely from scratch without relying on any pre-trained behavior policy or model. Specific approaches, such as preference-based reinforcement learning (PbRL) and rating-based reinforcement learning (RbRL) \citep{christiano2017deep, white2024rating}, typically train a classification model to predict human preferences or ratings on unseen behaviors generated by the agent. These preferences or ratings are then transformed into reward signals, often through normalization, to train the agent's policy. Specifically, PbRL utilizes binary human preferences over pairwise comparisons between two different segments generated by the agent under the current policy. The classification model is proposed to predict the probability that one segment is preferred over the other, and this model is then used to derive a reward function that aligns the agent’s behavior with human preferences. RbRL is developed upon PbRL by differentiating the quality of agent behaviors through discrete human ratings rather than binary preferences. Instead of comparing two segments, RbRL collects individual segments and asks humans to assign ratings based on the perceived quality. These ratings are then used to train a classifier that predicts the rating category for unseen behaviors, and the predicted class scores are normalized to serve as reward signals.

We argue that humans make rating decisions not solely by classifying different rating categories, but also by following certain criteria that are inherently regressive. For example, when assigning a score to a segment, humans not only choose between “good” or “bad” categories but also assess how far the given segment deviates from an internal quality threshold, effectively performing a form of ordinal regression. Therefore, we aim to develop a method that learns rewards from human ratings by incorporating both classification and regression. The authors in \cite{kendall2018multi} introduced a principled approach to multi-task learning by leveraging homoscedastic uncertainty to weigh the loss contributions of tasks such as semantic segmentation, instance segmentation, and depth regression. Their method addresses the challenge of manually tuning loss weights by learning task-dependent uncertainties directly from the data, leading to improved performance compared to training separate models. Inspired by this idea, we note that human rating decisions often involve both classification and regression-like reasoning. Accordingly, our method incorporates a multi-task learning framework to jointly learn from categorical and continuous aspects of human ratings, enabling more accurate reward inference in reinforcement learning settings.

\section{Preliminaries and Background}
\label{preliminaries}

In the standard RL setting, the RL agent interacts with the environment following a Markov Decision Process (MDP) which is defined by a tuple $(\mathcal{S}, \mathcal{A}, P, R, \gamma)$, where $\mathcal{S}$ is the state space, $\mathcal{A}$ is the action space, $P$ is the state transition probability distribution, $r$ represents the environmental rewards, and $\gamma \in [0, 1)$ is the discount factor that limits the influence of infinite future rewards. At each state $s \in \mathcal{S}$, the RL agent takes an action $a \in \mathcal{A}$, moves to the next state $s'$ determined by $P(s'|s,a)$. The environment provides a reward $r: \mathcal{S} \times \mathcal{A} \rightarrow \mathbb{R}$ at each interaction between itself and the RL agent. The goal is to learn a policy $\pi$ that maps states to actions to maximize the expected discounted cumulative rewards $\mathbb{E}[G] = \sum^{\infty}_{t=0}\gamma^tr_{t}$, where $t$ represents the current interaction step.


\subsection{Reward Learning from Human Ratings}

In the context of this paper, we consider a Markov Decision Process without environmental rewards $(MDP\backslash R)$, where the goal is to learn a reward function $\hat{r}$ from human ratings which will be used to train an optimal policy based on the cumulative reward $\hat{R}$. We follow the similar reward learning framework adopted in other works to learn $\hat{r}$ \citep{white2024rating}. Specifically, at each training step, a length-$k$ trajectory $\sigma = (s_0, a_0, ..., s_{k-1}, a_{k-1})$ is collected to be rated. The reward model $\hat{r}(s, a)$ is trained to predict returns over trajectory segments $\sigma$, where cumulative return $\hat{R}(\sigma)$ is normalized to $\tilde{R}(\sigma) \in [0, 1]$.

In the prior work \citep{white2024rating}, the reward predictor is updated via optimizing a multi-class cross-entropy loss function 
\begin{equation}\label{eq:rbrl_loss}
    L_{CE}(\hat{r}) = - \sum_{\sigma \in X}(\sum ^{n-1}_{i=0}\mu_{\sigma}(i)\log(Q_{\sigma}(i))),
\end{equation}
where $Q_{\sigma}(i)$ estimates the probability that the human assigns the segment $\sigma$ to the $i^{th}$ rating class. In particular, $Q_{\sigma}(i)$ is formulated as
\begin{equation}\label{eq:rbrl_Qfunc}
    Q_{\sigma}(i)= \frac{e^{-k(\tilde{R}(\sigma)-\bar{R}_i)(\tilde{R}(\sigma)-\bar{R}_{i + 1})}}{\sum_{j=0}^{n-1} e^{-k(\tilde{R}(\sigma)-\bar{R}_j)(\tilde{R}(\sigma)-\bar{R}_{j + 1})}}, 
\end{equation}
where $\bar{R}_i$ are rating boundaries derived from human label distribution to match the proportion of predicted ratings with that of human annotations. Essentially, $Q_{\sigma}(i)$ assigns a higher probability for a sample whose normalized cumulative return falls within the $i^{th}$ rating class with a lower bound $\bar{R}_i$ and an upper bound $\bar{R}_{i+1}$.

Although prior work has shown effectiveness in learning from human ratings using a classification-based framework, such an approach has inherent limitations when attempting to model the underlying reward structure implied by those ratings.
As shown in \eqref{eq:rbrl_loss}, the cross-entropy loss treats each rating class independently and penalizes only misclassifications, without considering the ordinal relationships between rating values. This can be suboptimal when human ratings carry scalar or continuous meaning. For instance, a rating of 3 should be considered closer to 4 than to 1, but cross-entropy fails to reflect such proximity in the loss.

To better leverage the scalar nature of human ratings, we explore an additional formulation by training the reward predictor in a regression setting. This approach allows the model to more directly capture the underlying value structure conveyed by the ratings. The specifics of this method are detailed in the next section.

\begin{figure*}
    \centering    \includegraphics[width=0.95\linewidth]{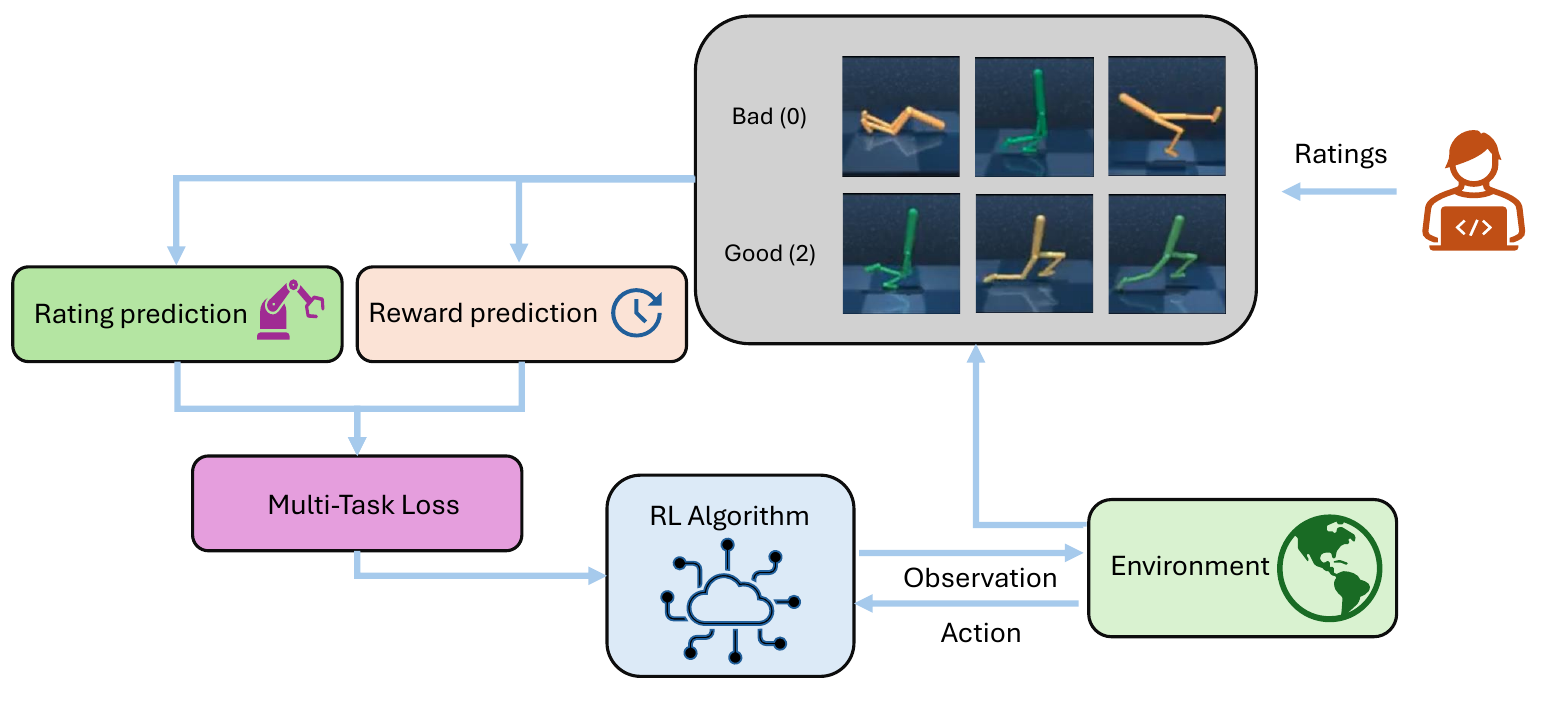}
    \caption{A schematic illustration of the proposed reward learning framework with human ratings.}
    \label{fig:flowchart}
\end{figure*}
\section{Technical Approach}
\label{techinical approach}

\subsection{Inferring Rewards from Human Ratings}
To enable the reward predictor to better interpret human ratings, we propose a novel rating-induced reward assignment mechanism given by
\begin{equation}\label{eq:log1p}
    R_i(\sigma) = \log(1 + \alpha \sum^{k}_{t=0}y_i),
\end{equation}

Where \( y_i \in n \) denotes the human-provided rating class, and \( \alpha > 0 \) is a scaling factor that adjusts the sensitivity of the reward to the rating magnitude. This transformation maps discrete rating values into a continuous and differentiable reward space. The logarithmic form compresses large rating values, ensuring numerical stability and promoting smoother gradient updates during training. This stability is particularly important because the model is trained from scratch using segments generated by a random policy, which results in a dataset dominated by low-rating classes and can otherwise lead to unstable or biased gradients.

With the proposed rating-induced reward assignment mechanism, the assigned reward can be treated as the ground-truth label for training the reward predictor on unseen segments. Specifically, we employ the standard mean squared error (MSE) loss \citep{schluchter2005mean} to minimize the discrepancy between the predicted cumulative rewards $\hat{R}(\sigma)$ and the assigned target rewards, which is formulated as
\begin{equation}\label{eq: L_reg}
    L_{reg} = (\hat{R}(\sigma) - R_i(\sigma))^2.
\end{equation}

We assume that human ratings are not solely derived from either classification or regression criteria, but rather reflect a comprehensive evaluation that integrates both aspects. Since the relative contributions of these aspects are not fixed, we model their uncertainty using Gaussian likelihoods with task-dependent (homoscedastic) variance \citep{kendall2018multi}. To capture this behavior, as illustrated in Figure \ref{fig:flowchart}, we utilize human ratings in both classification and regression forms to train a reward predictor that guides the RL policy. Specifically, we propose a novel loss function that jointly optimizes classification and regression objectives for training the reward predictor, enabling the reward predictor to learn from the complementary strengths of both paradigms. The multi-task loss function is designed as
\begin{equation}\label{eq:total loss}
    L = \frac{1}{2\lambda^2_{cls}}L_{CE}+\log\lambda_{cls}+\frac{1}{2\lambda^2_{reg}}L_{reg}+\log\lambda_{reg},
\end{equation}
where $\lambda_{cls}$ and $\lambda_{reg}$ are learnable uncertainty-based weighting factors representing the model's confidence in the classification and regression components, respectively.  
This novel loss function is derived from Gaussian likelihood for both regression and classification.
For regression, we assume output $y \in \mathbb{R}$ follows a Gaussian likelihood $p(y|\hat{R}(\sigma)) = N(y; \hat{R}(\sigma), \lambda_{reg}^2)$. The corresponding negative log-likelihood is $L_{reg} = -\log p(y|\hat{R}(\sigma)) = \frac{1}{2\lambda_{reg}^{2}}(y - \hat{R}(\sigma))^{2} + \frac{1}{2} \log(2\pi \lambda_{reg}^2)$. By ignoring constants, this simplifies to $L_{reg} = \frac{1}{2\lambda_{reg}^2}||y - \hat{R}(\sigma)||^2 + \log\lambda_{reg}$.
A similar ideal applies to the classification term, where we assume the classification output follows a scaled softmax likelihood. This leads to a negative log-likelihood of the form $L_{cls} = \frac{1}{2\lambda^2_{cls}}L_{CE}+\log\lambda_{cls}$.
This loss function allows the model to dynamically balance its emphasis between classification and regression, with lower uncertainty ($i.e.,$ smaller $\lambda$) resulting in higher weight for that component. 
$\log\lambda_{cls}$ and $\log\lambda_{reg}$ act as regularizers, preventing the model from trivially minimizing the overall loss by assigning excessively large values to $\lambda_{cls}$ and $\lambda_{reg}$, which would correspond to near-zero weights on the associated loss terms.

\begin{figure*}[!ht]
\centering
    \begin{subfigure}{0.3\textwidth}
        \centering
        \includegraphics[width=\textwidth]{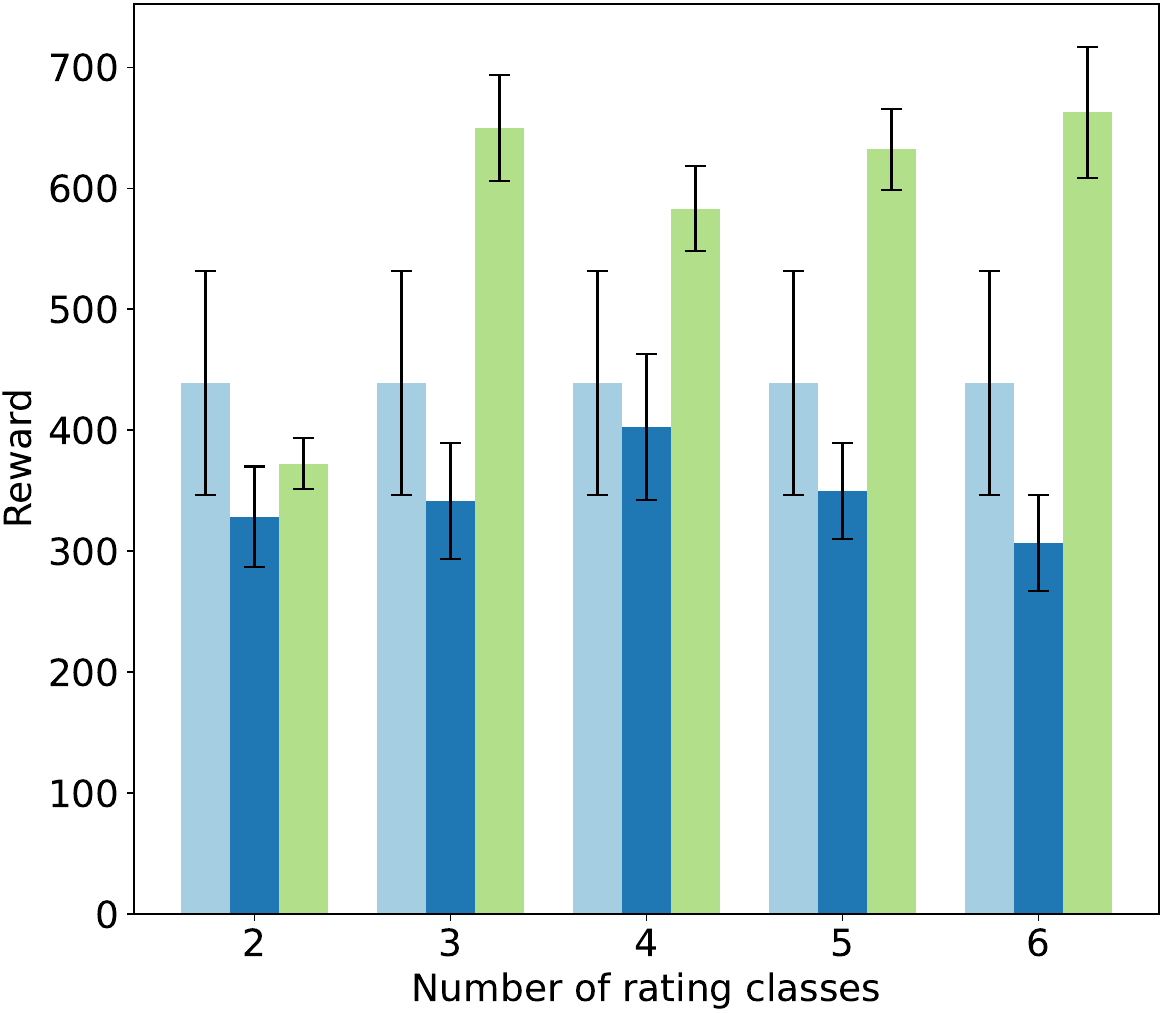}
        \caption{ {Cartpole-balance}}
        \label{fig:cartpole_balance}
    \end{subfigure}
    \hfill
    \begin{subfigure}{0.3\textwidth}
        \centering
        \includegraphics[width=\textwidth]{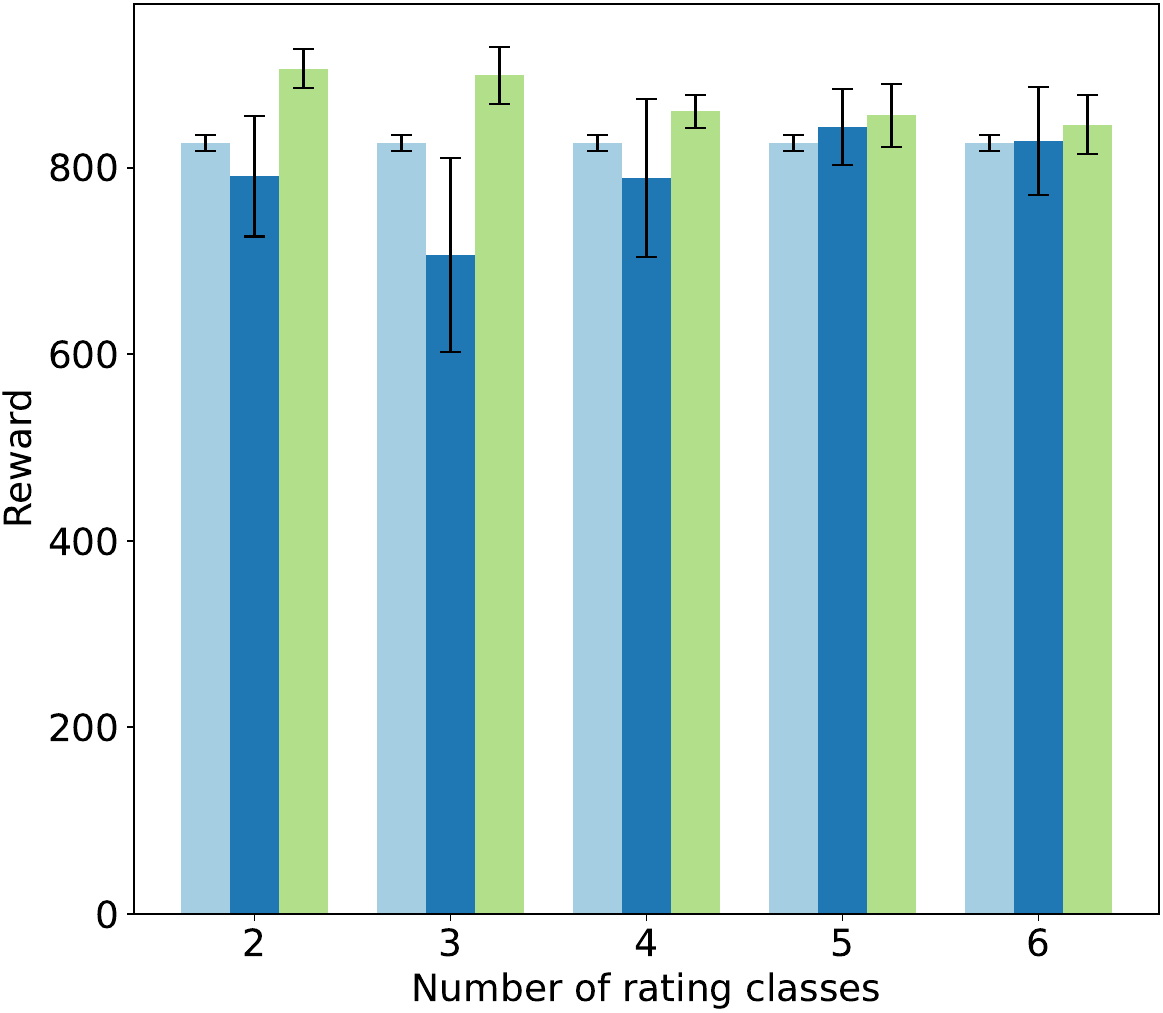}
        \caption{ {Ball-in-cup}}
        \label{fig:ball_in_cup}
    \end{subfigure}
    \hfill
    \begin{subfigure}{0.3\textwidth}
        \centering
        \includegraphics[width=\textwidth]{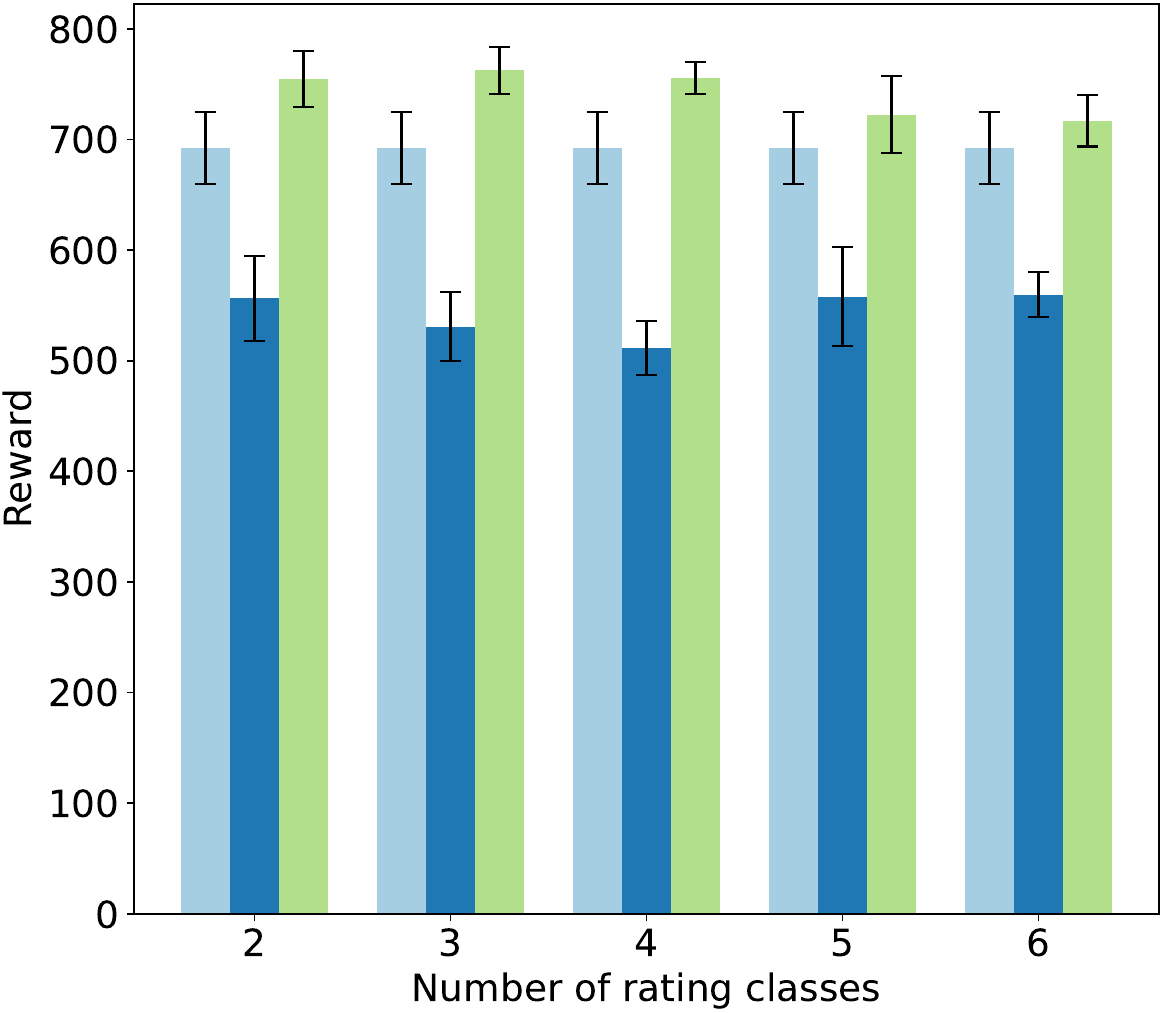}
        \caption{Finger-spin}
        \label{fig:finger_spin}
    \end{subfigure}
    \hfill
    \begin{subfigure}{0.3\textwidth}
        \centering
        \includegraphics[width=\textwidth]{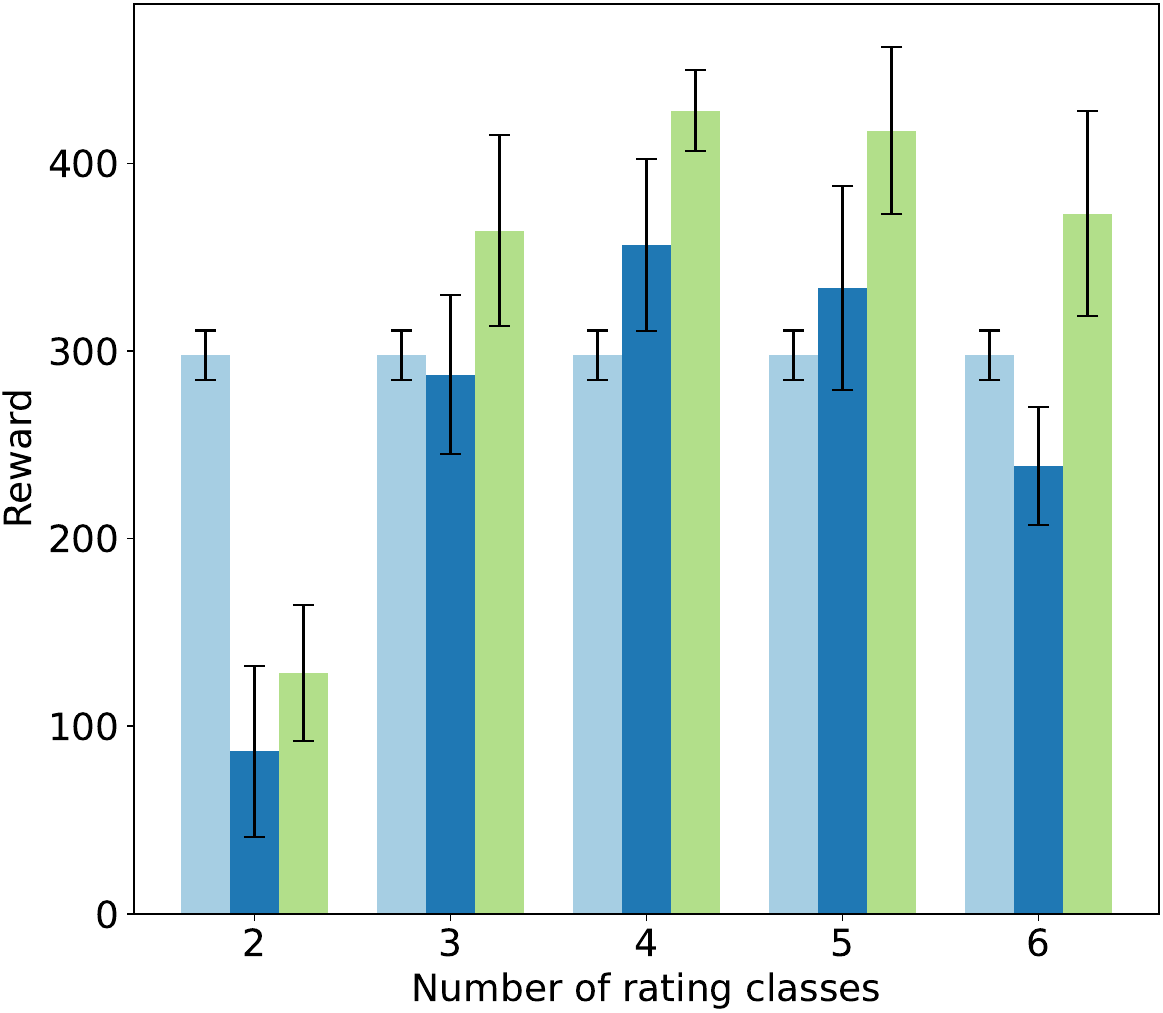}
        \caption{ {HalfCheetah}}
        \label{fig:half}
    \end{subfigure}
    \hfill
    \begin{subfigure}{0.3\textwidth}
        \centering
        \includegraphics[width=\textwidth]{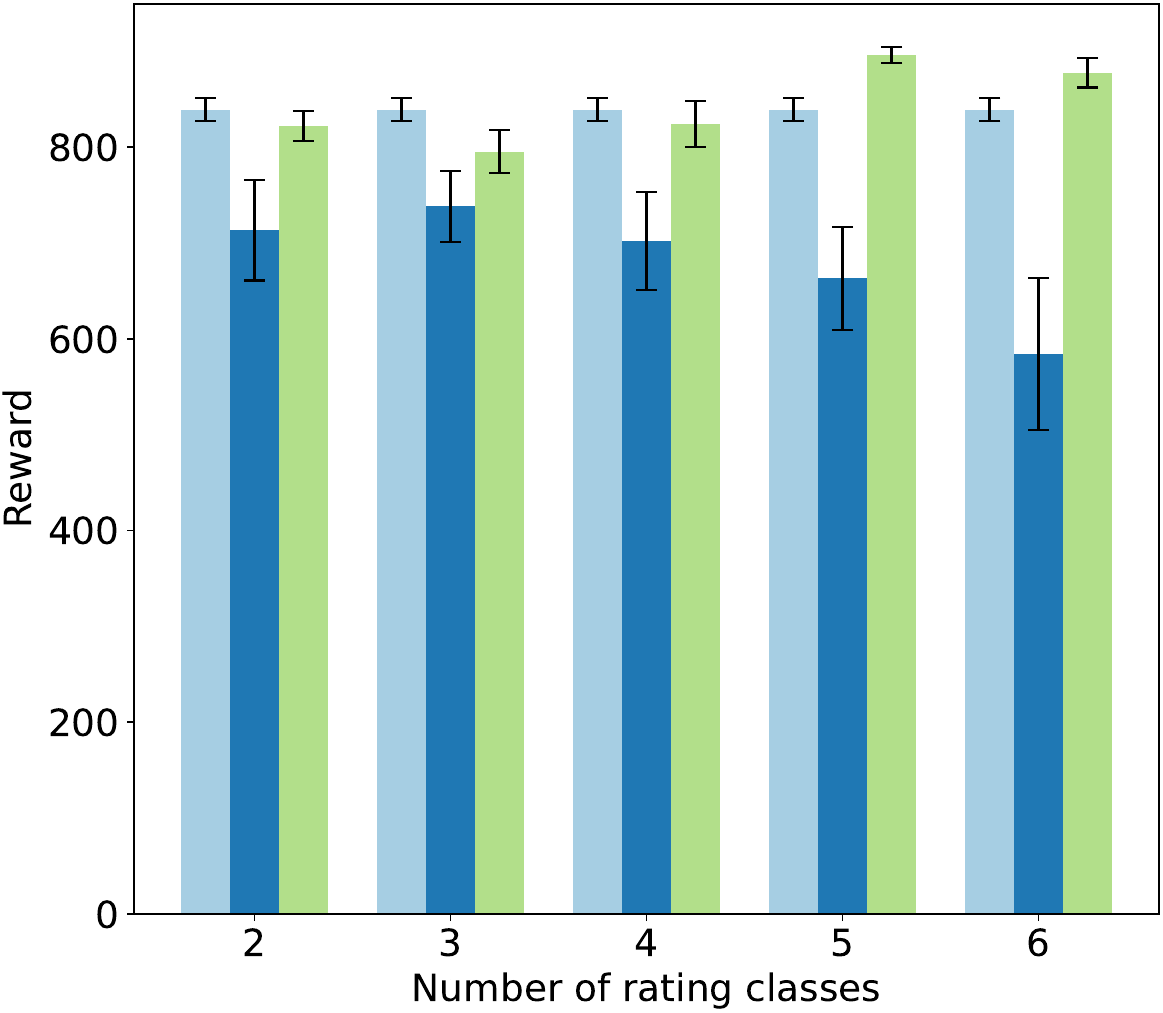}
        \caption{ {Walker}}
        \label{fig:walker}
    \end{subfigure}
    \hfill
    \begin{subfigure}{0.3\textwidth}
        \centering
        \includegraphics[width=\textwidth]{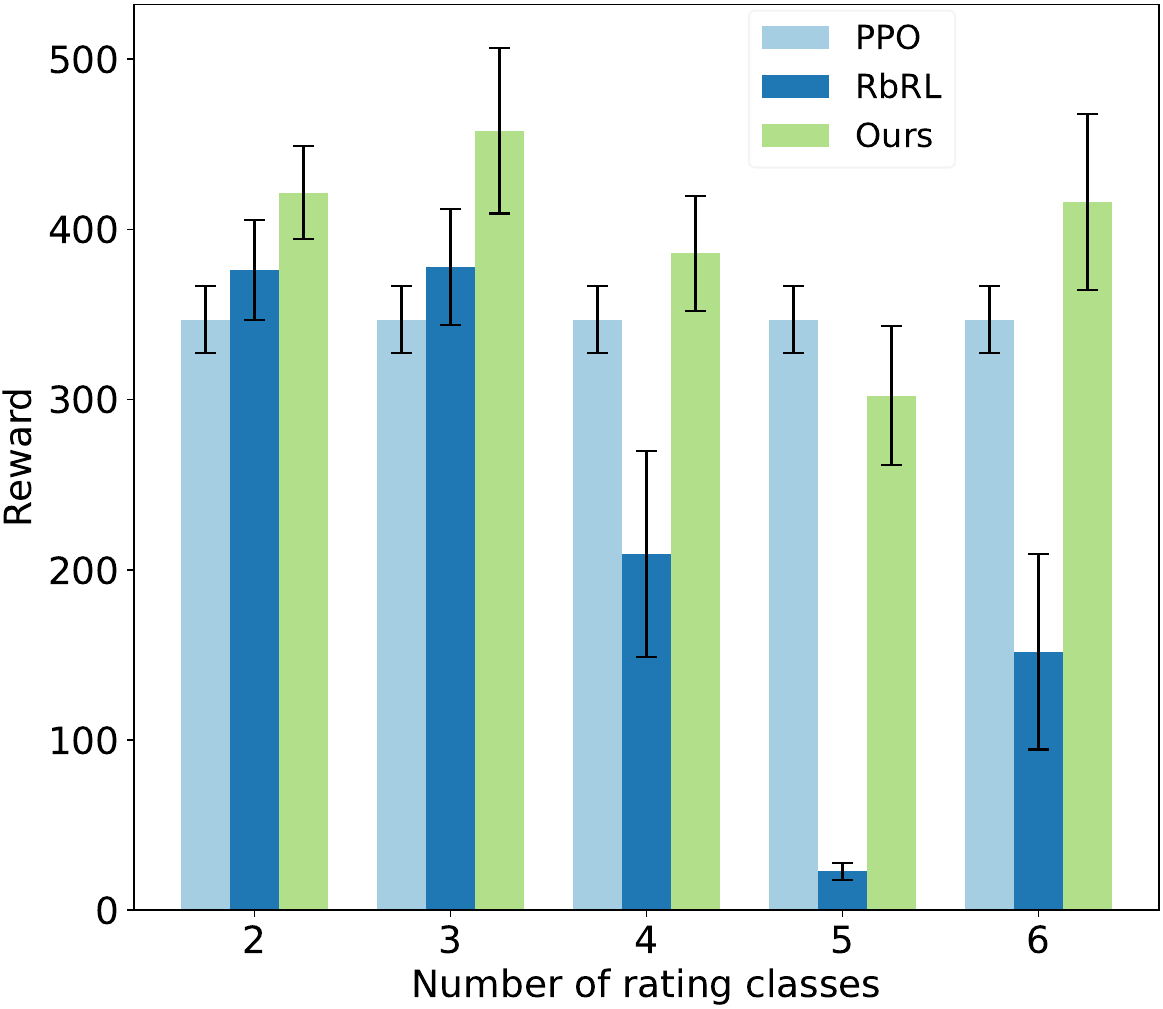}
        \caption{ {Quadruped}}
        \label{fig:quad}
    \end{subfigure}
    \caption{Empirical return comparison among different algorithms across six environments. The plots show the mean with the standard error over 10 runs.
}\label{fig:results}
\end{figure*}

\begin{figure*}[!ht]
\centering
    \begin{subfigure}{0.3\textwidth}
        \centering
        \includegraphics[width=\textwidth]{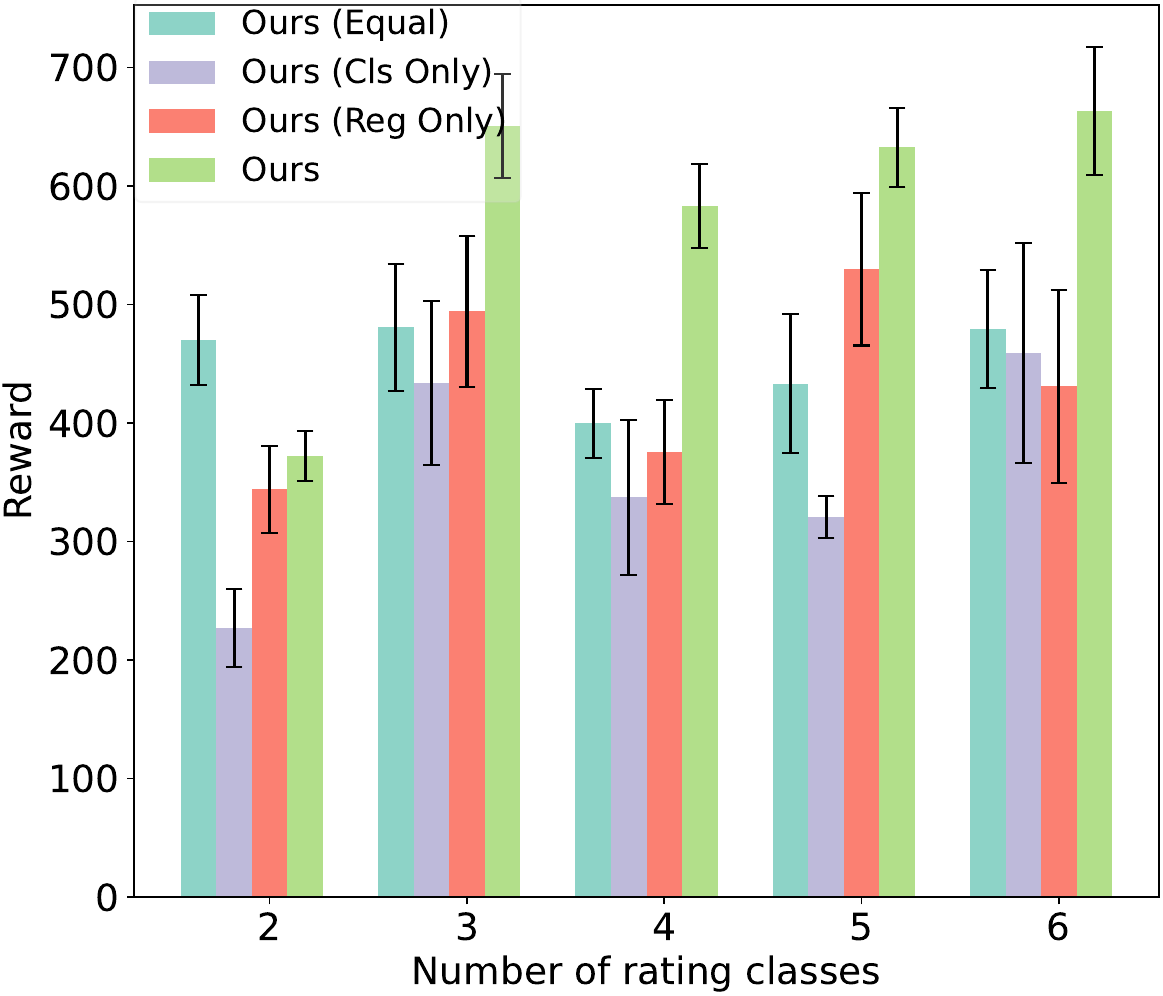}
        \caption{ {Cartpole-balance}}
        \label{fig:cartpole_balance}
    \end{subfigure}
    \hfill
    \begin{subfigure}{0.3\textwidth}
        \centering
        \includegraphics[width=\textwidth]{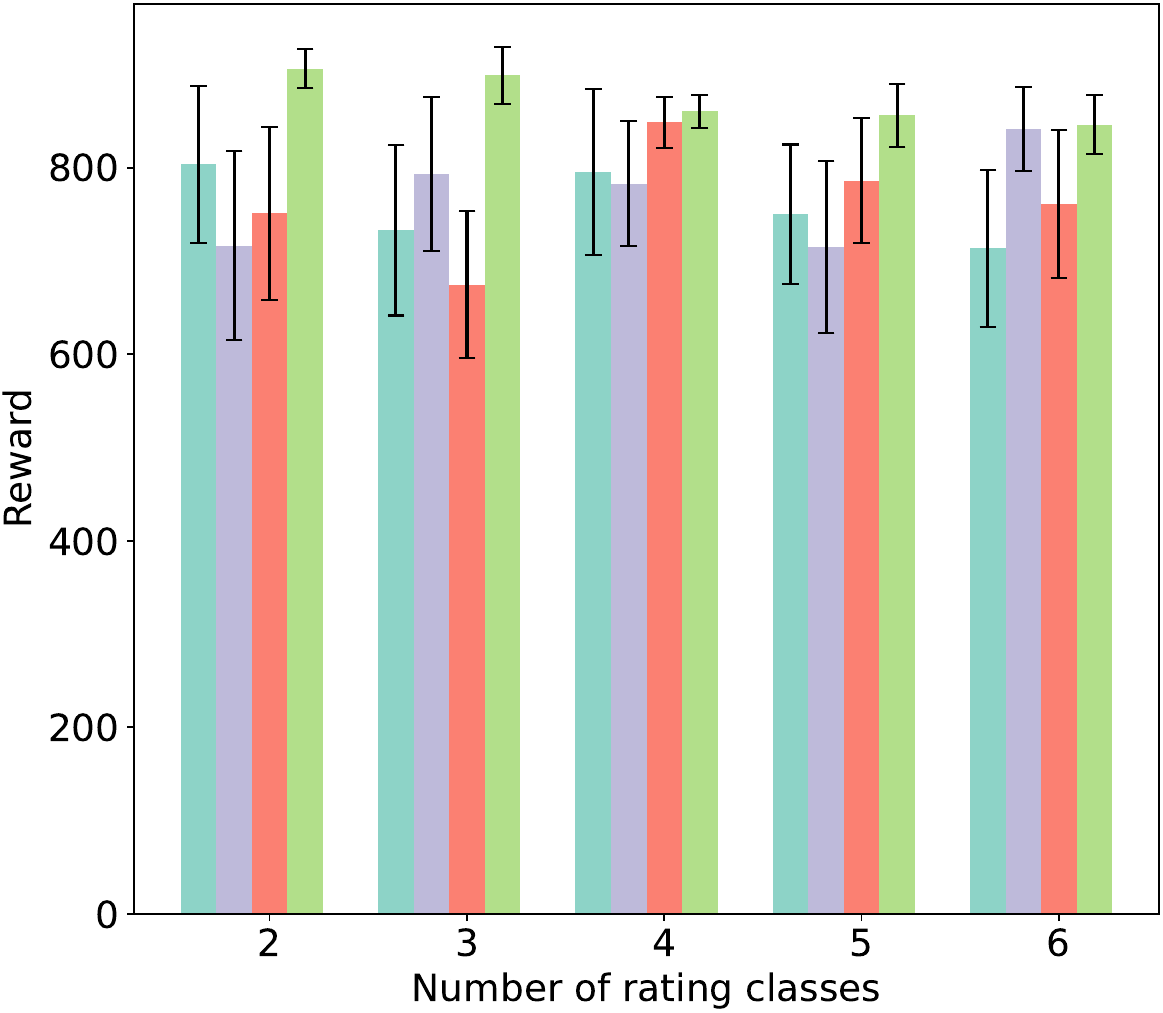}
        \caption{ {Ball-in-cup}}
        \label{fig:ball_in_cup}
    \end{subfigure}
    \hfill
    \begin{subfigure}{0.3\textwidth}
        \centering
        \includegraphics[width=\textwidth]{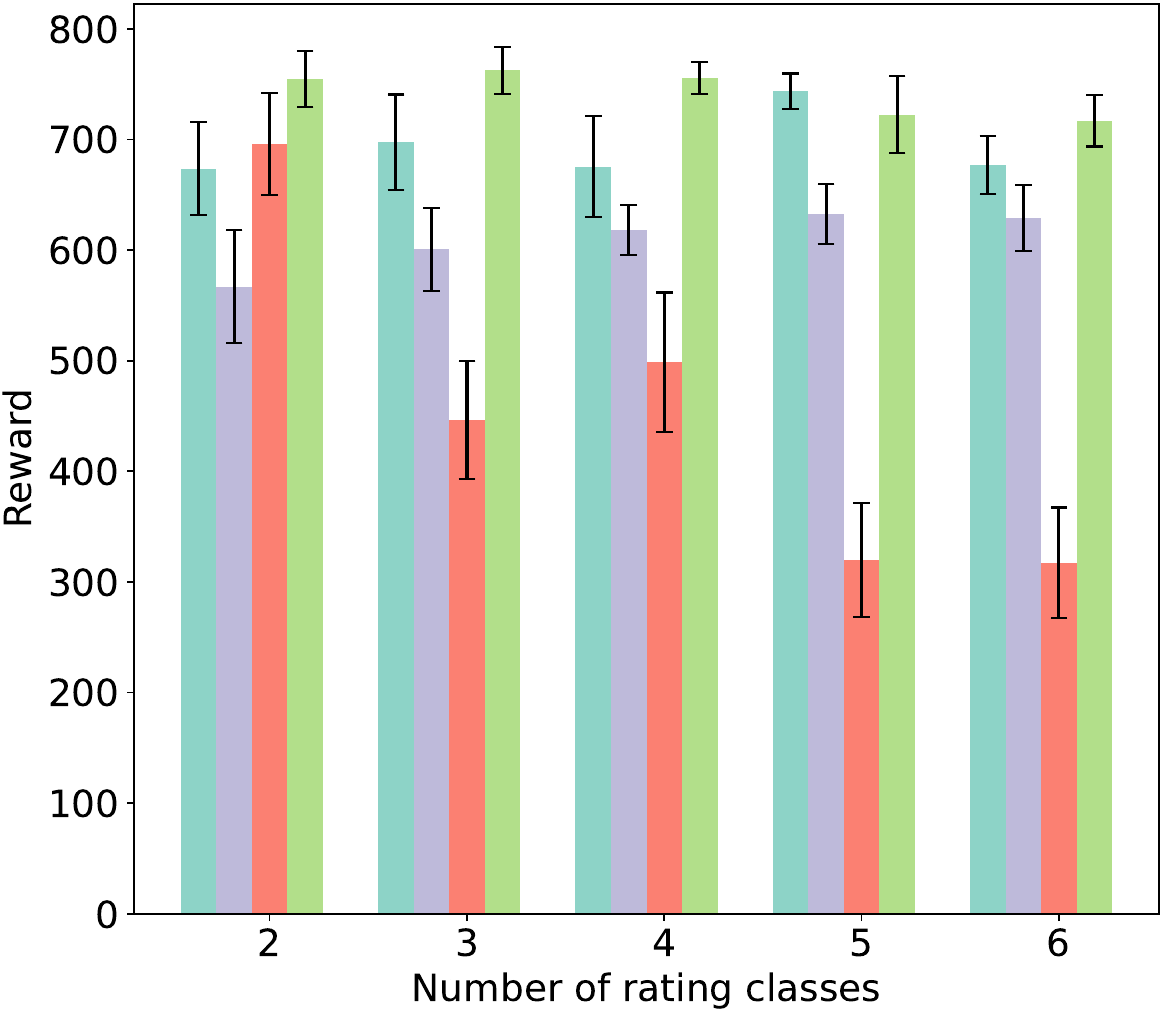}
        \caption{Finger-spin}
        \label{fig:finger_spin}
    \end{subfigure}
    \hfill
    \begin{subfigure}{0.3\textwidth}
        \centering
        \includegraphics[width=\textwidth]{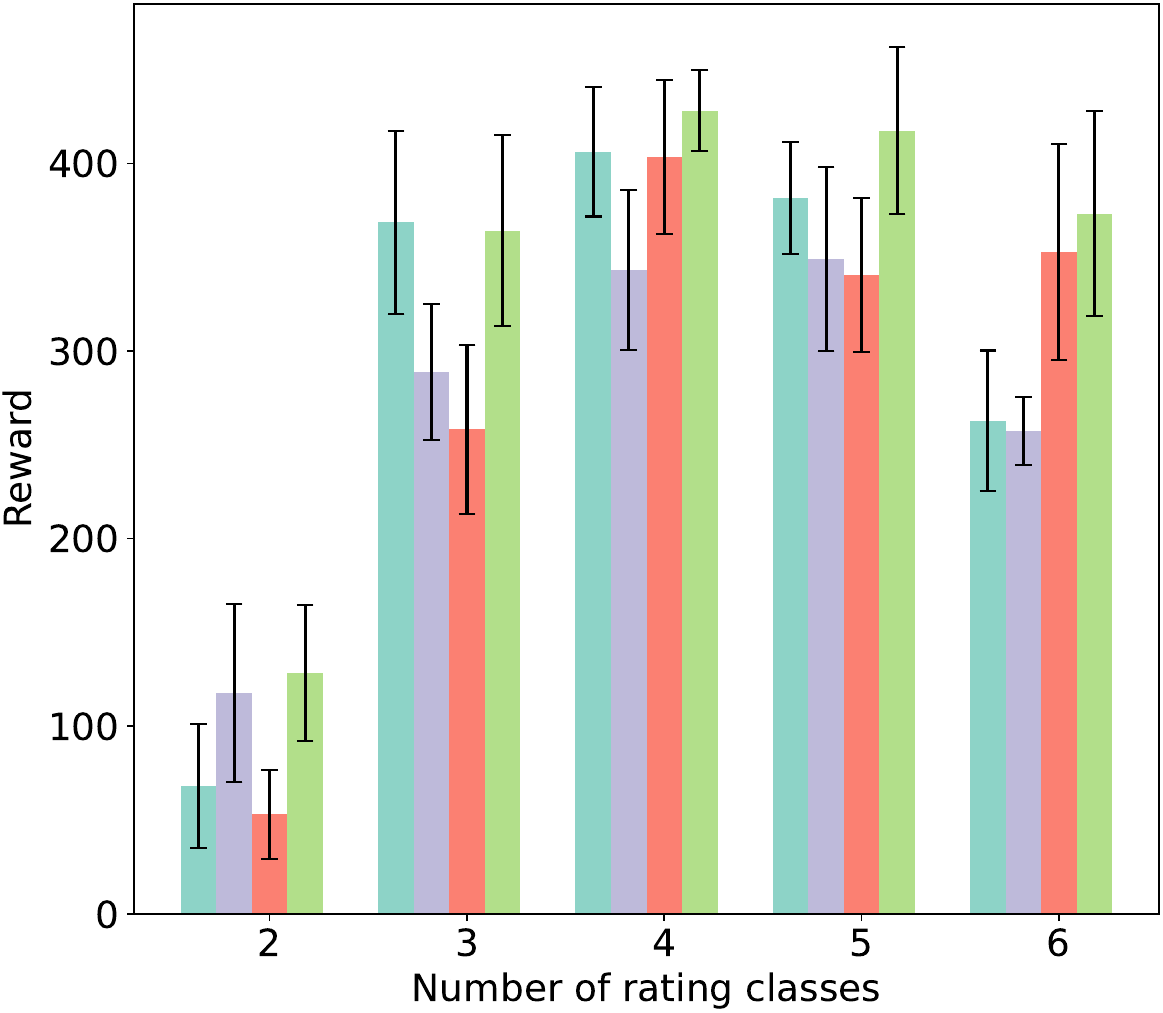}
        \caption{ {HalfCheetah}}
        \label{fig:half}
    \end{subfigure}
    \hfill
    \begin{subfigure}{0.3\textwidth}
        \centering
        \includegraphics[width=\textwidth]{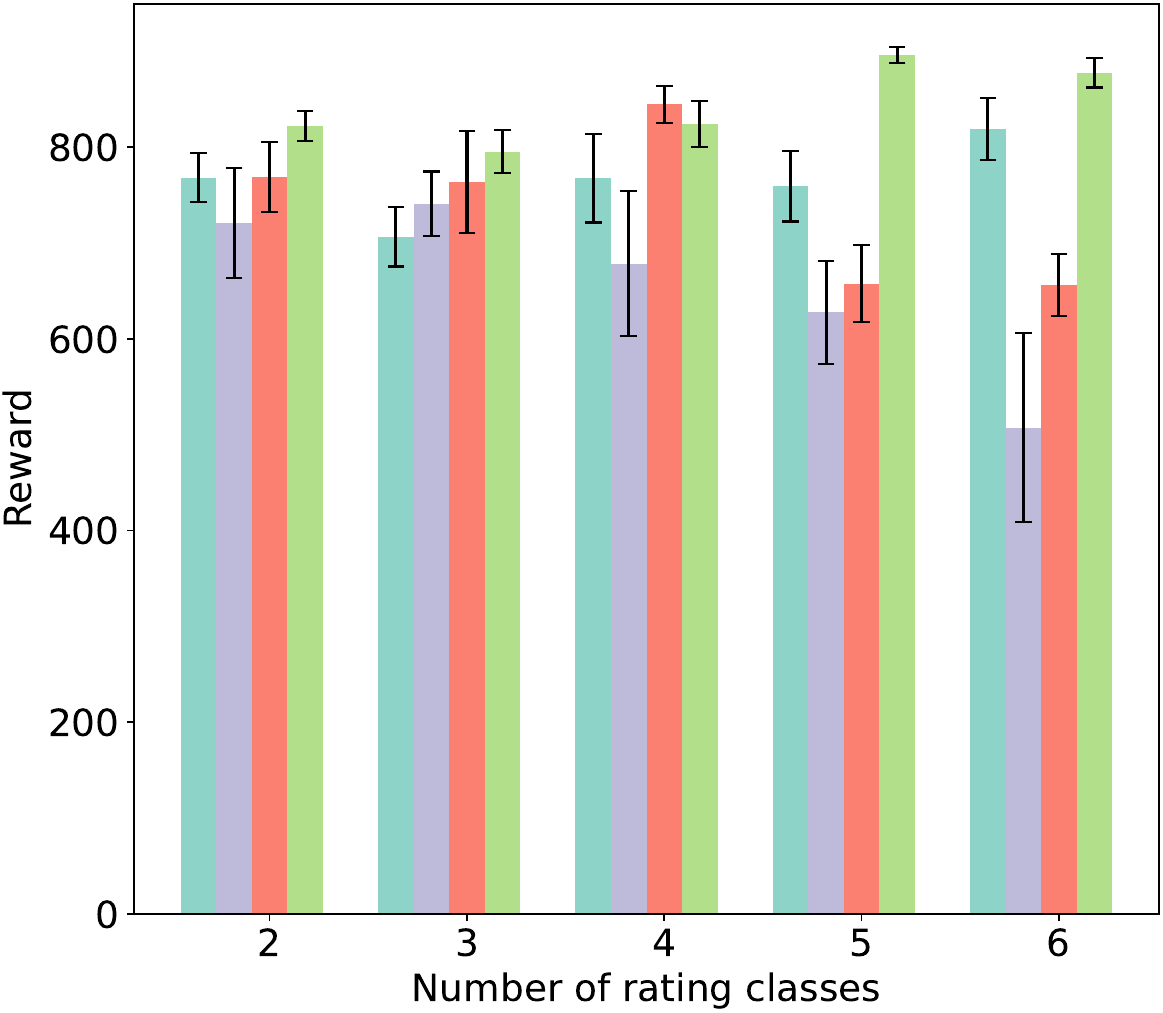}
        \caption{ {Walker}}
        \label{fig:walker}
    \end{subfigure}
    \hfill
    \begin{subfigure}{0.3\textwidth}
        \centering
        \includegraphics[width=\textwidth]{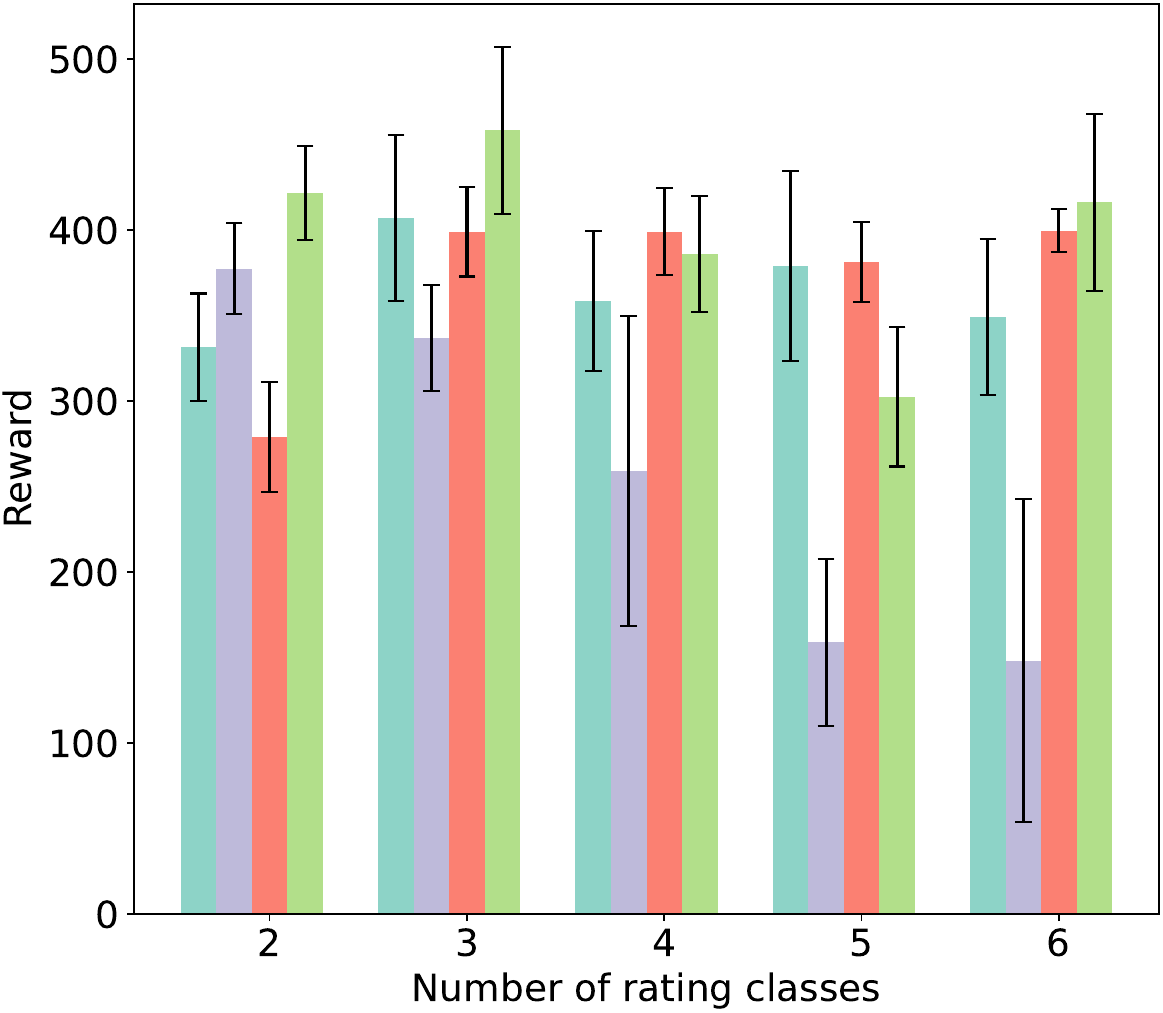}
        \caption{ {Quadruped}}
        \label{fig:quad}
    \end{subfigure}
    \caption{Ablation study comparing empirical returns across different configurations in six environments. The plots show the mean with the standard error over 10 runs.
}\label{fig:abl_results}
\end{figure*}

\section{Experiments and Results}
\label{experiemnts and results}
To evaluate the effectiveness of the proposed method, labeled as ``Ours", we compare it against several baselines across 6 DeepMind Control environments \cite{tassa2018deepmind}, namely, Cartpole-balance, Ball-in-cup, Finger-spin, HalfCheetah, Walker and Quadruped. These environments are characterized with continuous state and action spaces, and each varies in complexity. Specifically, Cartpole-balance is characterized with a simple 5-dimensional state space for cart and pole dynamics, and a 1-dimensional action space, representing discrete control forces. Ball-in-cup has an 8-dimensional state space capturing the relative positions and velocities of the involved objects, and a 2-dimensional action space controlling the cup's motion. Finger-spin is characterized with a 12-dimensional state space representing the finger and object dynamics, and a 2-dimensional action space controlling the finger's movement. HalfCheetah has a 17-dimensional state space, capturing joint positions, velocities, and body orientation, and a 6-dimensional action space, which corresponds to the torques applied to each joint. Walker has a 24-dimensional state space representing joint angles, velocities, and torso orientation, and a 6-dimensional action space, controlling forces applied to each limb. Quadruped is a more complex environment which characterized with a 78-dimensional state space, including joint positions, velocities, and full body orientation, along with a 12-dimensional action space, enabling torque-based control over each leg joint. Among these environments, Cartpole-balance is the simplest environment while Quadruped is the most complex one. 

In our experiments, we compare our proposed method against several baseline methods, including standard PPO \citep{schulman2017proximal} which updates policies upon environmental rewards (labeled as ``PPO"), RbRL \citep{white2024rating} which uses only a classification-based reward predictor (labeled as ``RbRL"), and a comprehensive ablation study. The ablations include (i) using equal weights for the classification and regression terms in the multi-task loss (labeled as ``Ours (Equal)"), namely a fixed weight 0.5 is assigned to both terms, and (ii) using only one of the loss terms, either the classification term or the regression term for policy updates, with uncertainty-based weighting (denoted as ``Ours (Cls Only)” for classification only and ``Ours (Reg Only)” for regression only), corresponding to the objectives $L = \frac{1}{2\lambda^2_{cls}}L_{CE}+\log\lambda_{cls}$ and $L = \frac{1}{2\lambda^2_{reg}}L_{reg}+\log\lambda_{reg}$, respectively.  We set $k = 30$ in \eqref{eq:rbrl_Qfunc} and $\alpha = 0.5$ in \eqref{eq:log1p} for cases involving classification, regression, or both.

To evaluate the effectiveness of our method across different rating classes, we consider configurations with 2, 3, 4, 5, and 6 rating classes. Each case is run for 4 million timesteps. The rating data are generated synthetically by manually defining the reward boundaries for each environment. For each configuration, we collect 2,000 synthetic ratings. All ratings are gathered during the early stage of the training process to train a reward function at an early phase. To ensure reproducibility, each configuration is evaluated over 10 independent runs, with performance reported as the average cumulative reward along with standard error. The specific hyperparameters used for each algorithm are detailed in Appendix \ref{sec:hyper}. 

As shown in Figure~\ref{fig:results}, our proposed method consistently outperforms RbRL, and exceeds PPO performance given a suitable number of rating classes. It is worth noting that even in reward-free environments, our method surpasses PPO across all environments by adapting to different numbers of rating classes, highlighting the flexibility and robustness of our approach. 

To further evaluate the effectiveness of our proposed method, we conduct an ablation study comparing mean returns and their associated standard errors under different configurations. As shown in Figure~\ref{fig:abl_results}, Ours consistently outperforms most other configurations across six environments. The only exceptions are in Cartpole-balance and Walker when $n = 2$ and $n = 4$, where Ours (Equal) slightly outperforms Ours, and Ours (Reg Only) marginally exceeds Ours. However, these differences are minimal. These results further highlight the adaptability of our proposed method. While Ours (Cls Only) or Ours (Reg Only) may perform well in isolated cases, they lack robustness across configurations. Ours Equal, which combines both objectives with fixed weights, generally outperforms its single-term counterparts, underscoring the importance of integrating both components. Overall, these findings suggest that dynamically learning the weights between classification and regression objectives is critical for achieving optimal performance across tasks.

\section{Conclusion}
\label{Conclusion}
In this work, we propose a multi-task learning framework that mimics how humans dynamically assign ratings. These ratings are leveraged to train a reward predictor by jointly optimizing classification and regression objectives. To capture the varying influence of these objectives, similar to how humans shift focus between discrete categories and continuous scales, we introduce learnable weights that adjust the contributions of classification and regression based on their confidence.
To evaluate the effectiveness of our approach, we conduct extensive experiments across multiple configurations and numbers of rating classes. The results demonstrate that dynamically balancing classification and regression leads to significant improvements over fixed-weight, single-objective baselines and standard PPO. This confirms the benefit of modeling the nuanced nature of human ratings in reward learning. 

Future work will explore the use of human ratings in more interactive or real-time settings to further evaluate the proposed method. Additionally, this approach has potential for real-world applications, such as robotics. We plan to test our algorithm on tasks involving path planning and trajectory following for robots ($e.g.,$ drones), especially in scenarios where designing a comprehensive reward function manually is challenging.




\section*{Impact Statement}

This paper presents a work whose goal is to advance the field of reinforcement learning from human feedback, specifically in reward learning from human ratings. By proposing a framework that dynamically integrates classification and regression to mimic human rating behavior, our method provides a more robust and flexible way to train agents in reward-free environments. This can enable the deployment of reinforcement learning in real-world domains where human feedback is more feasible than designing explicit reward functions, such as in robotics, education, or healthcare.

While the primary intention of our work is to improve learning efficiency and generalization through better use of human ratings, potential ethical considerations may arise depending on application domains. For instance, if human-generated ratings reflect unintended biases, these could be reinforced by the learning algorithm. Therefore, careful consideration of rating sources and fairness need to be considered when deploying such systems in sensitive settings.

Overall, we believe our contributions help advance the usability of reinforcement learning with human-in-the-loop settings, with both academic and practical benefits. No immediate negative societal impacts are anticipated, but the work may warrant further ethical review when applied to real-world decision-making tasks involving humans.

\vspace{-5 pt}
\section*{Acknowledgment}

This work was supported by the Office of Naval Research under Grant N000142212474 and the Army Research Office under Grant W911NF2310363.

\nocite{langley00}

\bibliography{example_paper}

\begin{thebibliography}{16}
\providecommand{\natexlab}[1]{#1}
\providecommand{\url}[1]{\texttt{#1}}
\expandafter\ifx\csname urlstyle\endcsname\relax
  \providecommand{\doi}[1]{doi: #1}\else
  \providecommand{\doi}{doi: \begingroup \urlstyle{rm}\Url}\fi

\bibitem[Christiano et~al.(2017)Christiano, Leike, Brown, Martic, Legg, and Amodei]{christiano2017deep}
Christiano, P.~F., Leike, J., Brown, T., Martic, M., Legg, S., and Amodei, D.
\newblock Deep reinforcement learning from human preferences.
\newblock \emph{Advances in neural information processing systems}, 30, 2017.

\bibitem[Dai et~al.(2023)Dai, Pan, Sun, Ji, Xu, Liu, Wang, and Yang]{dai2023safe}
Dai, J., Pan, X., Sun, R., Ji, J., Xu, X., Liu, M., Wang, Y., and Yang, Y.
\newblock Safe rlhf: Safe reinforcement learning from human feedback.
\newblock \emph{arXiv preprint arXiv:2310.12773}, 2023.

\bibitem[Kendall et~al.(2018)Kendall, Gal, and Cipolla]{kendall2018multi}
Kendall, A., Gal, Y., and Cipolla, R.
\newblock Multi-task learning using uncertainty to weigh losses for scene geometry and semantics.
\newblock In \emph{Proceedings of the IEEE conference on computer vision and pattern recognition}, pp.\  7482--7491, 2018.

\bibitem[Mnih et~al.(2013)Mnih, Kavukcuoglu, Silver, Graves, Antonoglou, Wierstra, and Riedmiller]{mnih2013playing}
Mnih, V., Kavukcuoglu, K., Silver, D., Graves, A., Antonoglou, I., Wierstra, D., and Riedmiller, M.
\newblock Playing atari with deep reinforcement learning.
\newblock \emph{arXiv preprint arXiv:1312.5602}, 2013.

\bibitem[Rose et~al.(2025)Rose, White, Wu, Lawhern, Waytowich, and Cao]{rose2025performance}
Rose, E., White, D., Wu, M., Lawhern, V., Waytowich, N.~R., and Cao, Y.
\newblock Performance optimization of ratings-based reinforcement learning.
\newblock \emph{arXiv preprint arXiv:2501.07755}, 2025.

\bibitem[Schluchter(2005)]{schluchter2005mean}
Schluchter, M.~D.
\newblock Mean square error.
\newblock \emph{Encyclopedia of Biostatistics}, 5, 2005.

\bibitem[Schulman et~al.(2017)Schulman, Wolski, Dhariwal, Radford, and Klimov]{schulman2017proximal}
Schulman, J., Wolski, F., Dhariwal, P., Radford, A., and Klimov, O.
\newblock Proximal policy optimization algorithms.
\newblock \emph{arXiv preprint arXiv:1707.06347}, 2017.

\bibitem[Sutton \& Barto(1998)Sutton and Barto]{sutton1998introduction}
Sutton, R.~S. and Barto, A.~G.
\newblock \emph{Reinforcement Learning: {A}n Introduction}.
\newblock The MIT Press, Cambridge, MA, 1998.

\bibitem[Sutton \& Barto(1999)Sutton and Barto]{sutton1999reinforcement}
Sutton, R.~S. and Barto, A.~G.
\newblock Reinforcement learning: An introduction.
\newblock \emph{Robotica}, 17\penalty0 (2):\penalty0 229--235, 1999.

\bibitem[Tassa et~al.(2018{\natexlab{a}})Tassa, Doron, Muldal, Erez, Li, Casas, Budden, Abdolmaleki, Merel, Lefrancq, et~al.]{tassa2018deepmind}
Tassa, Y., Doron, Y., Muldal, A., Erez, T., Li, Y., Casas, D. d.~L., Budden, D., Abdolmaleki, A., Merel, J., Lefrancq, A., et~al.
\newblock Deepmind control suite.
\newblock \emph{arXiv preprint arXiv:1801.00690}, 2018{\natexlab{a}}.

\bibitem[Tassa et~al.(2018{\natexlab{b}})Tassa, Doron, Muldal, Erez, Li, de~Las~Casas, Budden, Abdolmaleki, Merel, Lefrancq, Lillicrap, and Riedmiller]{tassa2018deepmindcontrolsuite}
Tassa, Y., Doron, Y., Muldal, A., Erez, T., Li, Y., de~Las~Casas, D., Budden, D., Abdolmaleki, A., Merel, J., Lefrancq, A., Lillicrap, T., and Riedmiller, M.
\newblock Deepmind control suite, 2018{\natexlab{b}}.
\newblock URL \url{https://arxiv.org/abs/1801.00690}.

\bibitem[Todorov et~al.(2012)Todorov, Erez, and Tassa]{todorov2012mujoco}
Todorov, E., Erez, T., and Tassa, Y.
\newblock Mujoco: A physics engine for model-based control.
\newblock In \emph{2012 IEEE/RSJ international conference on intelligent robots and systems}, pp.\  5026--5033. IEEE, 2012.

\bibitem[White et~al.(2024)White, Wu, Novoseller, Lawhern, Waytowich, and Cao]{white2024rating}
White, D., Wu, M., Novoseller, E., Lawhern, V.~J., Waytowich, N., and Cao, Y.
\newblock Rating-based reinforcement learning.
\newblock In \emph{Proceedings of the AAAI Conference on Artificial Intelligence}, volume~38, pp.\  10207--10215, 2024.

\bibitem[Wu et~al.(2023)Wu, Tao, and Cao]{wu2023value}
Wu, M., Tao, F., and Cao, Y.
\newblock Value of potential field in reward specification for robotic control via deep reinforcement learning.
\newblock In \emph{AIAA SCITECH 2023 Forum}, pp.\  0505, 2023.

\bibitem[Wu et~al.(2024)Wu, Siddique, Sinha, and Cao]{wu2024offline}
Wu, M., Siddique, U., Sinha, A., and Cao, Y.
\newblock Offline reinforcement learning with failure under sparse reward environments.
\newblock In \emph{2024 IEEE 3rd International Conference on Computing and Machine Intelligence (ICMI)}, pp.\  1--5. IEEE, 2024.

\bibitem[Wu et~al.(2025)Wu, White, Lawhern, Waytowich, and Cao]{wu2025rbrl2}
Wu, M., White, D., Lawhern, V., Waytowich, N.~R., and Cao, Y.
\newblock Rbrl2. 0: Integrated reward and policy learning for rating-based reinforcement learning.
\newblock \emph{arXiv preprint arXiv:2501.07502}, 2025.

\end{thebibliography}
\bibliographystyle{icml2025}

\newpage
\appendix
\onecolumn
\section{Hyperparameter Settings for Each Algorithm}

\label{sec:hyper}

This section presents the hyperparameters used for each algorithm evaluated in Section \ref{experiemnts and results}. Specifically, PPO is employed as the reinforcement learning backbone for RbRL, Ours (Equal), Ours (Cls Only), Ours (Reg Only), and Ours variants. The detailed hyperparameter settings are summarized in Table~\ref{tab: alg param}.

\begin{table*}[h!]\label{tab: alg parameters}
  \centering
  \begin{tabular}{c|c|c|c|c|c}
    \hline
    Algorithm & Clip param $\epsilon$ & Hidden Layers & Hidden Size & Activation Function & Learning Rate\\
    \hline
    PPO & 0.4 & 3 & 256 & $tanh$ & 0.00005\\
    RbRL & 0.4 & 3 & 256 & $tanh$ & 0.00005\\
    Ours (Equal) & 0.4 & 3 & 256 & $tanh$ & 0.00005\\
    Ours (Cls Only) & 0.4 & 3 & 256 & $tanh$ & 0.00005\\
    Ours (Reg Only) & 0.4 & 3 & 256 & $tanh$ & 0.00005\\
    Ours & 0.4 & 3 & 256 & $tanh$ & 0.00005\\
    \hline
  \end{tabular}
  \caption{Hyperparameter settings for all algorithms.}
  \label{tab: alg param}
\end{table*}



\end{document}